\documentclass[10pt,twocolumn,letterpaper]{article}

\usepackage{cvpr}
\usepackage{times}
\usepackage{epsfig}
\usepackage{graphicx}
\usepackage{amsmath}
\usepackage{amssymb}
\usepackage{diagbox}
\usepackage{multirow}

\usepackage{amsmath,amsfonts,bm}

\def\eqref#1{equation~\ref{#1}}

\def\1{\bm{1}}

\def\rx{{\textnormal{x}}}

\def\rz{{\textnormal{z}}}

\def\vz{{\bm{z}}}

\DeclareMathAlphabet{\mathsfit}{\encodingdefault}{\sfdefault}{m}{sl}
\SetMathAlphabet{\mathsfit}{bold}{\encodingdefault}{\sfdefault}{bx}{n}

\newcommand{\E}{\mathbb{E}}

\newcommand{\tool}{AdvSPADE\xspace}

\usepackage{subfigure}
\usepackage{algorithm}
\usepackage{algorithmic}
\usepackage{booktabs}

\usepackage[breaklinks=true,bookmarks=false]{hyperref}

\cvprfinalcopy 

\def\cvprPaperID{****} 

\begin{document}

\title{AdvSPADE: Realistic Unrestricted Attacks for Semantic Segmentation}

\author{Guangyu Shen\\
Purdue University\\
{\tt\small shen447@purdue.edu}
\and
Chengzhi Mao\\
Columbia University\\
{\tt\small cm3797@columbia.edu} 
\and
Junfeng Yang\\
Columbia University\\
{\tt\small junfeng@cs.columbia.edu}
\and
Baishakhi Ray\\
Columbia University\\
{\tt\small rayb@cs.columbia.edu}}

\maketitle



\begin{abstract}

Due to the inherent robustness of segmentation models, traditional norm-bounded attack methods show limited effect on such type of models. 
In this paper, we focus on generating unrestricted adversarial examples for semantic segmentation models.  
We demonstrate a simple and effective method to generate unrestricted adversarial examples using conditional generative adversarial networks (CGAN) without any hand-crafted metric. The na\"ive implementation of CGAN, however, yields inferior image quality and low attack success rate. Instead, we leverage the SPADE (Spatially-adaptive denormalization) structure with an additional loss item to generate effective adversarial attacks in a single step.
We validate our approach on the popular Cityscapes and ADE20K datasets, and demonstrate that our synthetic adversarial examples are not only realistic, but also improve the attack success rate by up to 41.0\% compared with the state of the art adversarial attack methods including PGD.
\end{abstract}
\section{Introduction}

\begin{figure*}[t]
\vspace{-5mm}
\centering
\includegraphics[width=0.9\linewidth]{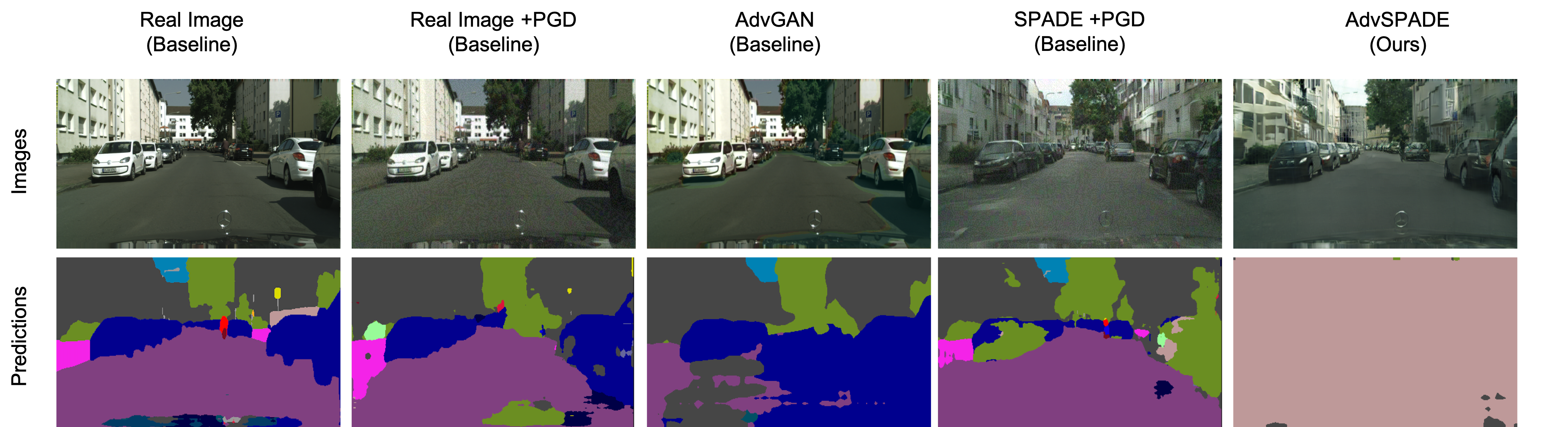}
\caption{\small{\textbf{Illustrating the effectiveness of  generated unrestricted adversarial example compared to norm bounded attacks}. The 1$^{st}$ column shows a real image from Cityscapes dataset and its prediction result of DRN-105 segmentation network. The 2$^{nd}$ and 3$^{rd}$ columns show the result of applying PGD and AdvGAN generated perturbation to real images, respectively. In the 4$^{th}$ column, PGD is applied to a synthesized image generated by SPADE, while the 5$^{th}$ column shows \tool's unrestricted adversarial image that completely fools DRN-105. Note that,  there are conspicuous noises in the 2$^{nd}$, 3$^{rd}$ and 4$^{th}$ column images. The segmentation results are still good for the 1$^{st}$, 2$^{nd}$ and 4$^{th}$ columns. While 3$^{rd}$ column shows better attack performance than traditional norm-bounded attacks, it is still worse than ours.  In contrast, in our case (5$^{th}$ column) the image is free from noise, it preserves the objects critical to driving (e.g., street lamps, cars, buildings),  with slight differences in colors and textures that do not affect the semantics for a human driver. However, the prediction of the segmentation model is totally wrong. This demonstrates the effectiveness of our unrestricted adversarial attacks for segmentation models.
}
}
\label{fig:fig0}
\vspace{-2mm}
\end{figure*}

Despite their impressive accuracy and wide adaption, 
deep learning (DL) models remain fragile to adversarial attacks~\cite{szegedy2013intriguing, CW, papernot2016transferability}, which raises serious concerns for deploying them into real-world applications,  especially in safety and security-critical systems. 

Extensive efforts have been made to combat these adversarial attacks:   robust models are trained such that they are not easily evaded by adversarial examples ~\cite{43405, Papernot_2016, madry2018towards}. 
Although these defense methods improve the models' robustness, they are mostly limited to addressing norm-bounded attacks such as PGD ~\cite{madry2018towards}, in which, human is not supposed to differentiate between original clean images and adversarial examples. 
However, two similar but non-identical images may also contain consistent semantic information from human perspective; e.g., a slightly rotated variant of an image is semantically similar to the original one~\cite{engstrom2017exploring}. 
Since such changes (e.g., lighting condition, certain textures, rotation, etc.) do not interfere with the human perception, 
a robust ML model should also remain invariant
to these realistic changes~\cite{tian2018deeptest}. 

However, such realistic variants of the clean images tend to 
have high attack success rate for robust models designed to defend norm-bounded attacks~\cite{song2018constructing}.
Thus, the realistic adversarial attacks beyond norm bound 
remain a major concern to those robust models,
which spur extensive efforts to explore stronger and realistic adversarial attacks, e.g., using Wasserstein bound measurement~\cite{wong2019wasserstein}, realistic image transformations~\cite{engstrom2017rotation} etc. 
In particular,~\cite{song2018constructing} propose unrestricted adversarial attacks using conditional GAN for the image classification models, a big step toward realistic attacks beyond human crafted constrains. 
However, due to their model design, they are mostly restricted to low-resolution images\textemdash for high resolution, the generated images are not very realistic.

The problem of achieving realistic adversarial attacks and defenses aggravate further for more difficult visual recognition tasks such as semantic segmentation, where one needs to attack order of magnitude more pixels while achieving a consistent human perception.  It is essential to make the segmentation models robust against adversarial attacks, especially due to their applicability in autonomous driving~\cite{ess2009segmentation}, medical imaging~\cite{Ronneberger_2015,shen2018brain}, and computer-aided diagnose system~\cite{milletari2016v}.  Unfortunately, we show that existing attack methods primarily designed for simple classification tasks do not generalize well to semantic segmentation. Besides, the inherent robustness of segmentation models~\cite{Arnab_2018} make existed adversarial attack methods less effective.
For instance, following the work of~\cite{Arnab_2018}, we show that for segmentation tasks the norm-bound perturbation becomes human visible since larger bounds are required for launching a successful attack. 
The unrestricted adversarial attack, on the other hand, is not constrained by the norm bounded budget, which can expose more vulnerabilities of a given machine learning model. However, the quality and resolution of the unrestricted adversarial images generated by those methods are low and limited to simple images like the handwritten digits. 

In this paper, we present the first realistic unrestricted adversarial attack, AdvSPADE,  for semantic segmentation models. Figure~\ref{fig:fig0} illustrates the effectiveness of our proposed method. To generate realistic images, 
we use SPADE~\cite{park2019semantic}, a state-of-the-art conditional generative model to generate high-resolution images (up to 1 million pixels). 
We then add an 
adversarial loss term on the original SPADE architecture to fool the target model. 
Thus, we create a wide variety of adversarial examples from a single image in a single step. 
Empirical results show that we can generate  realistic and more successful adversarial attacks than existing norm-bounded or GAN based attacks.
We further show that augmenting the training data with such realistic adversarial examples have the potential to improve the models' robustness. 

This paper makes the following contributions: 
(1) We propose a new realistic attack for semantic segmentation which defeats the existing state-of-the-art robust models. We demonstrate the existence of a rich variety of unrestricted adversarial examples besides the previously known ones. 
(2) We demonstrate that augmenting the training dataset with our new adversarial examples have the potential of improving the robustness of existing models. 
(3) We present an empirical evaluation of our approach using two popular semantic segmentation dataset. First, we evaluate the quality of our generated adversarial examples using Amazon Mechanical Turk and demonstrate that our samples are indistinguishable to natural images for humans. Second, we further show that our adversarial samples improve the attack success rate up to $41\%$.

\section{Related Work}
\label{gen_inst}

\noindent
\textbf{Semantic Segmentation.}
It can be considered as a multi-output classification task that provides more fine-granular information in the prediction ~\cite{Barrow:1981:ILD:3015381.3015385}.
Although plenty of network architectures have been proposed to address semantic segmentation task efficiently ~\cite{Ronneberger_2015,long2015fully,chen2017deeplab,badrinarayanan2017segnet}, 
very few studies~\cite{Arnab_2018,Xie_2017} look into the robustness of this class of networks against adversarial examples. 



\noindent
\textbf{Adversarial Attacks.}
Adversarial examples are carefully crafted to mislead the DL models' predictions, while still perceived identical or similar semantics with original images to the human. 
Researchers have proposed multiple methods for generating adversarial examples for the image classification tasks~\cite{43405,kurakin2016adversarial, CW, madry2018towards}, where the target model is fooled by the adversarial images.
Hand-crafted metrics, such as $L_p$ norm bound ~\cite{madry2018towards, Feinman2017DetectingAS} and Wesstrasien distance ~\cite{wong2019wasserstein}, are applied to the generation process to preserve the semantic meaning of the adversarial examples to human. 

Generative Adversarial Network ($GAN$), popular for image manipulation ~\cite{yao20183daware,NIPS2018_8240,hong2018learning,yu2019free}, image-to-image translation~\cite{Isola_2017,Wang_2018,park2019semantic}, etc., was also leveraged to generate adversarial examples. 
Xiao et al.~\cite{Xiao_2018} proposed $AdvGAN$ to generate norm-bounded perturbation for the classification task. Song et al.~\cite{song2018constructing} used $GAN$ to generate {\em unrestricted adversarial attacks} for image classification. By leveraging Auxiliary Classifier Generative Adversarial Network ($AC$-$GAN$)~\cite{Odena2016ConditionalIS}, the model was able to generate low-quality adversarial examples from scratch and beyond any norm bound.~\cite{wang2019atgan} further proposed a new generative model, $AT$-$GAN$, to learn a transformation between a pre-trained $GAN$ and an adversarial $GAN$. Concurrent to our work, \cite{dunn2019adaptive} leveraged $GAN$ to generate adaptive unrestricted examples for classification.
Compared to the restricted image manipulation or local image editing attacks, one of the key features of these unrestricted attacks is that it can generate more realistic images semantically similar (may not be identical) to original benign images. Thus, these images can fool even the robust ML models that can withstand norm-bounded restricted adversarial examples. 

However,  these $GAN$-based attack methods mainly focus on classification tasks, and as we will show in Section~\ref{sec:result}, do not generalize well for segmentation tasks. 
The one closest to us, Song et al.~\cite{song2018constructing}, adopt two-step procedures: first train $AC$-$GAN$ to generate benign images and then optimize an adversarial loss objective function w.r.t the input noise $z$ of $AC$-$GAN$. In this setting, the variety of unrestricted adversarial examples is restricted by the dimension of $z$, usually fixed to a small number (10,100,256). Thus, they tend not to work well in high-resolution datasets.  In contrast, we combine the image generation and unrestricted adversarial image generation into a single stage by adding an adversarial loss item in its objective function, and optimize them together. 
This leads to a better attack success rate and a larger mIoU drop than Song's method, as shown in Section~\ref{sec:result}.

A few studies focused on the adversarial attack on modern semantic segmentation networks.
~\cite{Arnab_2018} conducted the first systematic analysis about the effect of multiple adversarial attack methods on different modern semantic segmentation network architectures across two large-scale datasets. ~\cite{Xie_2017} propose a new attack method called Dense Adversary Generation $(DAG)$, which generates a group of adversarial examples for a bunch of state-of-the-art segmentation and detection deep networks. However, all of the attack methods rely on norm-bounded  perturbations, which only cover a small fraction of all the feasible adversarial examples.~\cite{cisse2017houdini} further  advances the restricted attacks using surrogate loss functions for semantic segmentation to make non-differential IOU loss to be approximate differential, however, due to the expensive computationally, traditional norm-bounded attack methods such as $PGD$, $FGSM$ are more widely used in practise.

\noindent
\textbf{Defense Methods.}
Adversarial training is the state-of-the-art method for training robust classifiers~\cite{43405,Lyu_2015,Shaham2018UnderstandingAT,szegedy2013intriguing,44873,papernot2017extending,Xu_2018,madry2018towards}. Besides,~\cite{Arnab_2018} evaluated other defense methods like the input transformation,  including rescaling, JPEG compression, Gaussian blur, HSV jitter, grayscale against adversarial attack on semantic segmentation networks. These input transformation methods, however, were shown to rely on obfuscated gradients and give a false sense of robustness ~\cite{athalye2018obfuscated}. Thus, 
~\cite{athalye2018obfuscated} endorsed the robustness of the model trained with adversarial training.
In this paper, we demonstrate that the robustness can be improved by training with examples generated by our method. 


\section{Generating Adversarial Examples}
\label{headings}

\begin{figure*}[!ht]
\vspace{-10mm}
\begin{center}
\includegraphics[width=0.8\linewidth]{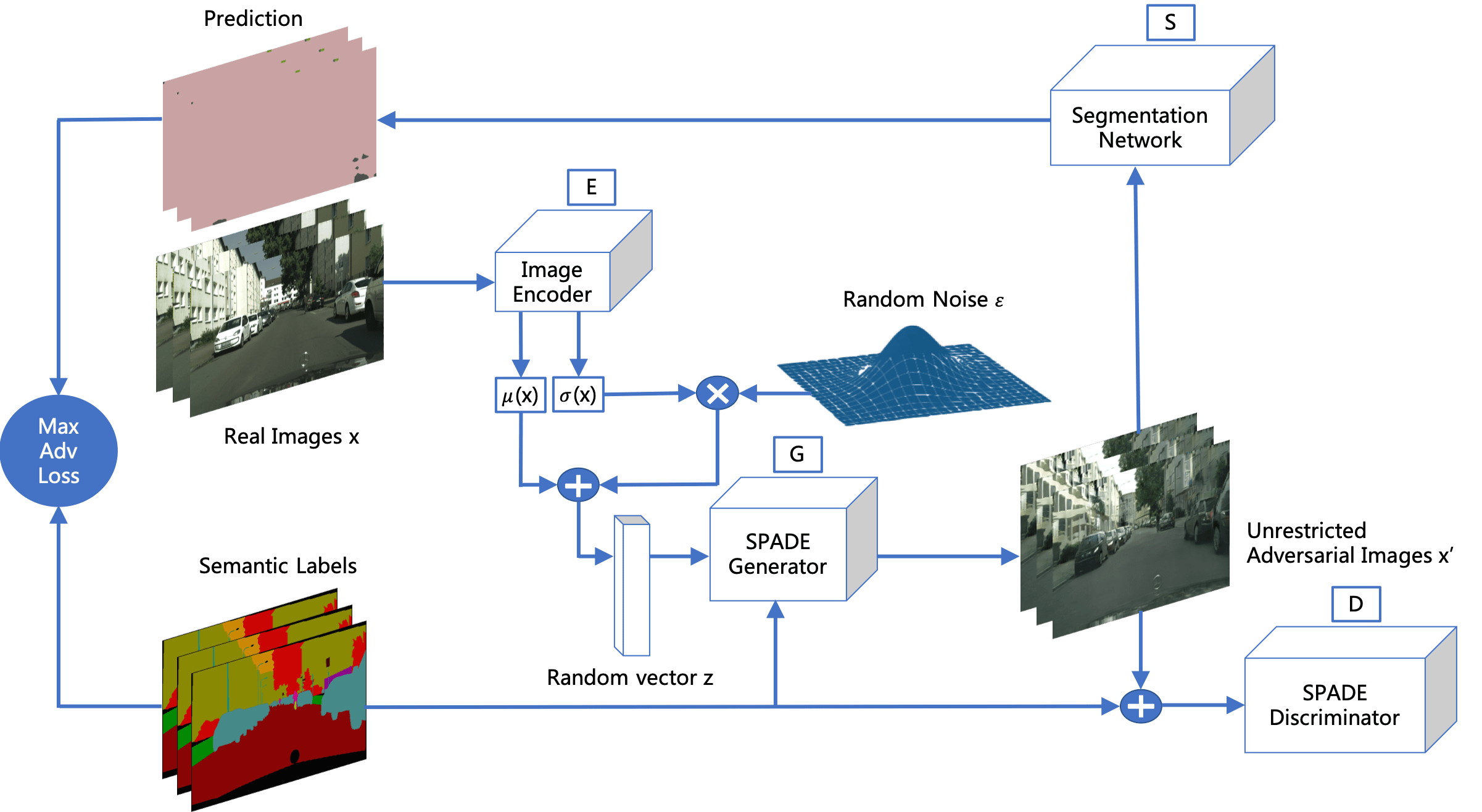}
\end{center}
\caption{\small{\textbf{Illustration of proposed AdvSPADE Architecture for generating unrestricted adversarial examples.} Image encoder $E$ takes real images $x$ as input to compute mean and variance vectors ($\mu(x), \sigma(x)$) and apply reparameterization trick to generate random noise $z$. SPADE generator $G$ considers $z$ and semantic labels $s$ and generates synthetic images $x^\prime$. Next, $x^\prime$s are fed into a fixed pre-trained target segmentation network $S$ and encouraged to mislead $S$'s predictions by maximizing adversarial loss between predictions and semantic labels. Meanwhile, $x^\prime$, as SPADE discriminator $D$'s input, also aims to fool $D$. $D$ is trained to reliably distinguish between generated ($x^\prime$) and real images $(x)$. Random sampled $\epsilon$ brings randomness into the model so that $G$ can generate various adversarial examples. Notice that, due to the adversarial loss item, prediction results at the top left corner of the figure are completely mis-segmented.}}
\label{fig:fig1}
\vspace{-5mm}
\end{figure*}

We will now introduce our methodology for generating unrestricted adversarial examples for semantic segmentation. For this purpose, we leverage a conditional Generative Adversarial Networks, SPADE. The main goal of a standard conditional GAN is to synthesize realistic images that will fool the discriminator. The generation of adversarial attack, however, also requires to fool the segmentation model under attack.  Hence, we add an additional loss function to fool both the discriminator and the segmentation model.  Figure~\ref{fig:fig1} shows the overall workflow. The rest of the section describes the steps 
in details.

\noindent
\textbf{Unrestricted Adversarial Examples.}~
Consider $\mathcal I$ as a set of images and $\mathcal C $ be the set of all possible categories for $\mathcal I$.  Suppose $o:\mathcal O \subseteq \mathcal I \rightarrow \mathcal C$ is an oracle that can map any image from its domain $\mathcal O$ to $\mathcal C$ correctly. A classification model $\mathcal F: \mathcal I \rightarrow \mathcal C$ can also provide a class prediction for any given images in $\mathcal I$. Under the assumption that $\mathcal F \neq o$, an unrestricted adversarial example $ x$ is any image which meets following requirements ~\cite{song2018constructing}: $x \subseteq \mathcal O$, $o(x) \neq \mathcal F(x)$.

\noindent
\textbf{Conditional Generative Adversarial Networks.} 
 A Conditional Generative Adversarial Network ~\cite{mirza2014conditional} consists of a generator G and a discriminator D and they are both conditioned on auxiliary information $y$. Combining random noise $z$ and extra information $y$ as input, G is able to map it to a realistic image. The discriminator aims to distinguish the real images and synthetic images from the Generator. G and D correspond to a minimax two-player game and can be formalized as $\min_G\max_DV(G,D) = \E_{\rx\sim P_{data}(x)}[\log D(x|y)] + \E_{\rz\sim P_{z}(z)}[\log (1-D(G(z|y)))]$ 


\noindent
{\bf Unrestricted Adversarial Loss.}
We design an adversarial loss term for the unrestricted adversarial examples generation. We mainly focus on the untargeted attack in this paper though our approach is general and can be simply applied on targeted attack. Intuitively, the SPADE generator is trained to mislead the prediction of target segmentation network. The synthetic images are not only required to fool the discriminator for the conditional GAN but also need to be mis-segmented by the target segmentation network. To achieve this goal, we introduce the target segmentation network into the training phase and aim to maximize the loss of the segmentation model while keeping the quality and semantic meaning of the synthetic images. We denote the target segmentation network by $S$, SPADE generator by $G$, input semantic label  by $y$, and the input random vector by $z$. We define the untargeted version of Unrestricted Adversarial Loss as follows:
\begin{equation}
\setlength{\abovedisplayskip}{0pt}
\setlength{\belowdisplayskip}{0pt}
\displaystyle
L_{ATK} =  - \E_{\rz\sim P_{z}(z)} \log f(S(G(z|y)),y)
\end{equation}


We select Dice Loss ~\cite{Sudre_2017} as the objective function $f$. An image encoder $E$ processes a real image $I$ and generates a mean vector $\mu_{E(I)}$ and a variance vector $\sigma_{E(I)}$ and then compute the noise input $z$ according to reparameterization trick ~\cite{kingma2013autoencoding}.
\begin{equation}
\setlength{\abovedisplayskip}{0pt}
\setlength{\belowdisplayskip}{0pt}
\displaystyle
\vz = \mu_{E(I)} + \sigma_{E(I)}\cdot \epsilon ,  \epsilon \sim \mathcal N(0,I)
\end{equation}

The complete objective function of AdvSPADE then can be written as:
\begin{align} \label{G_loss}
\setlength{\abovedisplayskip}{0pt}
\setlength{\belowdisplayskip}{0pt}
\displaystyle
\min \limits_{G,E}((\max \limits_{D_1,D_2,D_3}\sum_{k=1,2,3}\mathcal L_{GAN}(G,D_k)) \notag
\\ +\lambda_0 \sum_{k=1,2,3}{\mathcal L_{FM}(G,D_k)}) +\lambda_1 \mathcal L_{VGG}(G) \notag
\\+ \lambda_2 \mathcal L_{KLD}(E) +\lambda_3 \mathcal L_{ATK}(S,G))
\end{align}

We follow the definition of feature matching loss $\mathcal L_{FM}$ and perceptual loss $\mathcal L_{VGG}$ in~\cite{Wang_2018}. The feature matching loss, $\mathcal L_{FM}(G,D_k) = \E_{(y,x)} \sum_{i=1}^{T} \frac{1}{N_i}[||D^{(i)}_k(y,x) - D^{(i)}_k(s,G(y))||_1]$, is able to stabilizes the GAN training by minimizing the distance between features of synthesis and real images from multiple layers of the discriminator. 
The VGG perceptual loss ($\mathcal L_{VGG}(G) = \sum_{(i=1)}^N \frac{1}{M_i}[||F^{(i)}(x) - F^{(i)}(G(y))||_1]$) plays a similar role as $\mathcal L_{FM}$ 
by introducing a pretrained VGG network.  
For $\mathcal L_{KLD}$, we borrow the definition from ~\cite{park2019semantic}. $\mathcal L_{KLD} = D_{KL}(q(z|x)||p(z))$. It calculates the KL divergence between a prior distribution $p(z)$, a standard Gaussian distribution and a variational distribution $q(z|x)$.  To speed up the generation process as well as the quality of synthesized images, we follow Spatially-adaptive denormalization, as proposed by ~\cite{park2019semantic}. 

\noindent
{\bf Spatially-adaptive denormalization}. Our model uses SPADE architecture ~\cite{park2019semantic} as the conditional GAN model, where the Batch Normalization ~\cite{ioffe2015batch} is replaced with Spatially-adaptive denormalization. This method is proved to maintain the semantic segmentation information which will get lost during the subsampling. Please refer to supplementary materials for details.

\section{Experimental Set-up}
\label{eval}

\noindent
\textbf{Datasets.}
We evaluate \tool on two large image segmentation datasets: Citsycapes~\cite{Cordts2016Cityscapes} and 
ADE20K~\cite{zhou2016semantic}. 
Cityscapes contains street view images from $50$ German cities and $19$ semantic classes, and it consists of $3000$ training and $500$ validation images. ADE20K covers $150$ semantic classes in multiple real world scenes, where the training and validation set contains $20100$ and $2000$ images respectively.

\noindent
\textbf{Training.}
Following~\cite{park2019semantic}, we apply the Spectral Norm~\cite{miyato2018spectral} in all layers of generator and discriminator. We train \tool with $50$ epochs on Cityscapes. For ADE20K, we run $100$ epochs (rather than $200$ epochs reported in~\cite{park2019semantic} due to  ADE20K's large size and computation limits). 
We set the learning rate of the generator and discriminator both equal to $0.0002$ and start to decay learning rate linearly from $50^{th}$ epoch when trained on ADE20K. We employ the ADAM~\cite{kingma2014adam} with $\beta_1 = 0.5$, $\beta_2 = 0.999$. In Equation {\ref{G_loss}}, we set $\lambda_0 = 10$, $\lambda_1 = 10$, $\lambda_2 = 0.05$ for both Cityscapes and ADE20K and $\lambda_3 = 10$ for Cityscapes, $\lambda_3 = 70$ for ADE20K respectively. All experiments are done on a single NVIDIA TITAN Xp GPU.

\noindent
\textbf{Baseline Models.} We compare \tool generated attacks with traditional \textit{norm-bounded} attacks in two settings: (i) real images with  perturbation and (ii) GAN-generated clean images with perturbation. For (ii), we generate clean images with vanilla SPADE and then add norm-bounded perturbation over the synthetic images. 
For a better comparison, we choose the same segmentation networks as target networks for each dataset as~\cite{park2019semantic} mentioned: DRN-D-105~\cite{yu2017dilated} for Cityscapes, Upernet-101 for ADE20K~\cite{xiao2018unified}. Besides, we also select several state-of-the-art open source segmentation networks to evaluate the transferability of \tool as a black box setting: 
DRN-38, DRN-22~\cite{yu2017dilated,Yu2016}, DeepLab-V3~\cite{chen2017rethinking}, 
PSPNet-34-8s~\cite{Zhao_2017} for Cityscapes, PPM-18, MobilenetV2, Upernet-50, PPM-101~\cite{zhou2018semantic} for ADE20K.

We also compare AdvSPADE with two GAN-based attacks ~\cite{Xiao_2018,song2018constructing} on Cityscapes.
For AdvGAN, we adapt the GAN structure of~\cite{Xiao_2018} for the segmentation task. First, we change the classification target network $f$ to the segmentation target network DRN-105 for Cityscapes. Second, we change the loss function of adversarial loss items from cross-entropy to dice loss for a fair comparison with ours. We reproduce their method with two different hyper-parameter settings ($\lambda_{adv} = -1,-100$)  
(see Table~\ref{advGAN}).

We implement Song et al.'s~\cite{song2018constructing} method on SPADE. We first train a SPADE to generate clean images and fix SPADE parameters. Second, we apply their designed objective function, substitute target classification network $f$ to DRN-105, and change the loss function of adversarial loss item from cross-entropy to dice loss. We set the dimension of input random noise $z$ to 256. We run SPADE with 50 epochs and then run another 50 epochs to optimize adversarial loss on Cityscapes. We also test the transferability of Song et al.'s attack on a black-box setting (see Table~\ref{mIoU on Cityscapes and ADE20K}).


\begin{figure*}[t]
\begin{center}
\includegraphics[width=0.9 \linewidth,height = 200 pt]{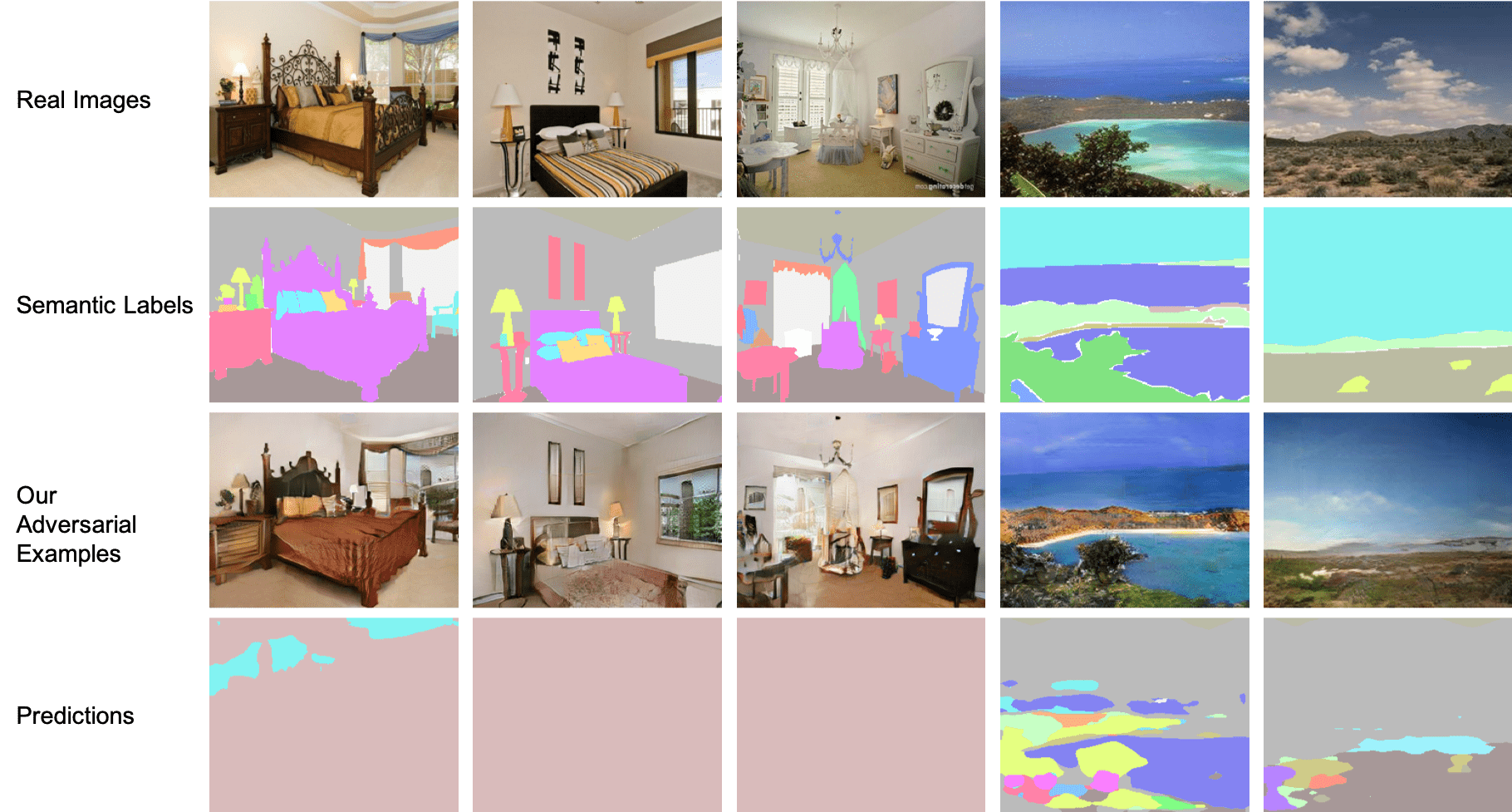}
\end{center}
\caption{\small{\textbf{Visual Results on ADE20K.} 
As we can see, the semantic meaning of the generated adversarial examples are well aligned to the original image, but are different in the style and mis-predicted by the segmentation model. For example, the color and strip on {bed} in the first two columns are changed, but human still perceive them as bed while the segmentation model predict the wrong label. The results demonstrate the effectiveness of our method for generating realistic adversarial examples that mislead the target model.}}
\label{fig:fig4}
\end{figure*}

\noindent
\textbf{Evaluation metric.}
Due to the dense output property of the semantic segmentation task, the evaluation of the attack success rate is different from that of the classification ~\cite{song2018constructing}. 
Let $\mathcal I ^{H \times W \times C} $ be a set of RGB images with height $H$ and width $W$ and channel $C$. Let $\mathcal L^{H \times W} $ be the set of semantic labels for the corresponding images from $\mathcal I$.  Suppose $o:\mathcal O \subseteq  \mathcal I \rightarrow \mathcal L$ is an oracle that can map any images from its domain $\mathcal O$ which presents all images that look realistic to humans to $\mathcal L$ correctly. A segmentation model $\mathcal S: \mathcal I \rightarrow \mathcal L$ can provide pixel-wise predictions for any given images in $\mathcal I$. 
We evaluate the following two categories of adversarial examples:
1) Given a constant $\epsilon > 0$ and a hand-craft norm $\left\lVert \cdot \right\rVert$, a \textit{restricted adversarial example} $x$ is an image that meets the following conditions:  $x \subseteq \mathcal O$, $\exists x^\prime  \subseteq \mathcal O$ $\left\lVert x-x^\prime \right\rVert < \epsilon$, $\frac{\sum_{i,j}( o_{i,j}(x)  \neq \mathcal S_{i,j}(x))}{H \cdot W} > \theta $.
(2) An \textit{unrestricted adversarial example} $x$ is an image that meets following requirement: $x \subseteq \mathcal O$, $\frac{\sum_{i,j}( o_{i,j}(x)  \neq \mathcal S_{i,j}(x))}{H \cdot W} > \theta $.
Here, $o_{i,j}(x),\mathcal S_{i,j}(x)$ stands for the prediction given by oracle $o$ and segmentation network $\mathcal S$ at pixel $x(i,j)$ respectively. $\theta \in [0,1] $ is a hyperparameter that is set to $\theta = 0.95$. 

Given the nature of semantic segmentation task, misclassifying a single pixel does not lead an image to fall into the class of adversarial examples. A legitimate adversarial example should have the property that the majority of pixels in it are misclassified (measured by mIoU score), and the adversarial image still looks realistic to humans (measured by FID score) with the same semantic meaning as the original images (measured by Amazon Turk). In particular, we use following three measures: 
1.~{\bf Mean Intersection-over-Union (mIoU)}: 
    For measuring the effect of different attack methods on the target networks, we measure the drop in recognition accuracy using mIoU  score which is widely used in semantic segmentation tasks ~\cite{Cordts2016Cityscapes,zhou2016semantic}\textemdash lower mIoU score means better adversarial example. 
2. {\bf Fr\'{e}chet Inception Distance (FID)}: We use FID ~\cite{heusel2017gans} to compute the distance between the distribution of our adversarial examples and the distribution of the real images; small FID stands for the high quality of generated images. 
3.~\textbf{Amazon Mechanical Turk (AMT)}: AMT is used to verify the success of our unrestricted adversarial attack. 
    Here, we randomly select $250$ generated adversarial images under two experimental settings from each dataset to generate AMT assignments. Each assignment is answered by $3$ different workers and each worker has $3$ minutes to make decision. We use the result of a majority vote as each assignment's final answer.

\section{Experiment Result}
\label{sec:result}

\begin{table*}[t]
\caption{\small{\textbf{The effectiveness of our proposed attack (\tool) under mIoU metric under both whitebox
and transfer based blackbox attacks.} 
Lower mIoU means better attack. 
Results show that \tool successfully misleads the segmentation models while both the real and synthetic images are predicted correctly by the models.
}
}
\label{mIoU on Cityscapes and ADE20K}
\centering
\scriptsize
\setlength{\tabcolsep}{3pt}
\begin{tabular}{l|lrrrr|lrrr}
\toprule
& \multicolumn{5}{c|}{Cityscapes} &    \multicolumn{4}{c}{ADE20K}\\
\midrule
Attacks & Seg Model  & Real Images  & SPADE  & Song et al~\cite{song2018constructing}  & \tool 
& Seg Model & Real Images & SPADE  & \tool\\
\midrule
Whitebox & DRN-105 & 0.756 & 0.620 & 0.461 & 0.010 & Upernet-101   & 0.420    & 0.403 & 0.011 \\
\midrule
Transfer-   & DRN-38      &0.714  &0.551      &0.520 &0.407   
&MobilenetV2 &0.348  &0.317      &0.110  \\

based & DRN-22      &0.68   &0.526      &0.489  &0.387          &PPM-18     &0.340  &0.362      &0.102 \\
        
blackbox &  DeepLab-V3  &0.68   &0.54   &0.501 &0.425                &Upernet-50 &0.404  &0.395  &0.096 \\
                          
& PSPNet-34-8s &0.691 &0.529  &0.495 &0.441               &PPM-101     &0.422 &0.409  &0.078\\
\bottomrule
\end{tabular}
\vspace{-5mm}
\end{table*}

\begin{table}[ht]
\caption{\small{\textbf{FID Comparison between our AdvSPADE and state-of-art semantic image synthesis models}. The results show that AdvSPADE outperforms Pix2PixHD and CRN and achieve comparable FID with vanilla SPADE on Cityscapes.}}
\label{FID}
\centering
\scriptsize
\setlength{\tabcolsep}{3pt}
\begin{tabular}{l|rrrr}
\toprule
\diagbox{Dataset}{Model} &Vanilla SPADE &Pix2PixHD  &CRN  &\bf{\tool} \\
\midrule
Cityscapes          &62.939     &95.0       &104.7      &\bf{67.302}\\
ADE20K              &33.9       &81.8       &73.3       &\bf{53.49}\\
\bottomrule
\end{tabular}
\vspace{-5mm}
\end{table}

\noindent
\textbf{Evaluating Generated Adversarial Images.}
Here, we compare the adversarial images generated from the original real images and the clean synthetic images created by vanilla SPADE using mIoU (Table \ref{mIoU on Cityscapes and ADE20K}) and FID scores (Table~\ref{FID}).  
Table~\ref{mIoU on Cityscapes and ADE20K} shows that compared to vanilla SPADE, generated images under whitebox attack can lead to a giant decline on mIoU score (from $0.62$ to $0.01$ for DRN-105, from $0.403$ to $0.011$ for Upernet-101). On different network architectures, our adversarial examples can also decrease mIoU 
(around $20\%$ on Cityscapes, $30\%$ on ADE20K) showing strong transferability of our examples across models. 

Compare to vanilla SPADE, the FID of our adversarial examples increases slightly ($62.939$ to $67.302$ on Cityscapes, $33.9$ to $53.49$ on ADE20K, see Table~\ref{FID}) indicating our samples have comparable quality and variety. Note that we only train \tool half epochs as reported in~\cite{park2019semantic} and achieve $53.49$ FID on ADE20K, still smaller than other leading semantic image synthetic models. 
Figure~\ref{fig:fig4} shows qualitative results.
Moreover, by introducing an image encoder and KL Divergence loss, we can generate multi-modal stylized adversarial examples which are shown in the supplementary materials.



\begin{table*}[t]
\caption{\small{\textbf{Attack Success Rate under white-box setting} for our \tool and norm-bounded attacks (FGSM,PGD) with bound sizes: $0.25,1,8,32$. 
\tool achieves significantly higher attack success rates on both datasets. 
}}
\label{Success Rate}
\centering
\scriptsize
\begin{tabular}{lrrr|lrrr}
\toprule
\multicolumn{4}{c|}{DRN-105} & \multicolumn{4}{c}{Upernet-101} \\
\midrule
  &Bound Size      &Real Images  &Vanilla SPADE &  &Bound Size     &Real Images  &Vanilla SPADE\\
Method  &($\epsilon$)      &+Perturbation  &+Perturbation &Method  &($\epsilon$)      &+Perturbation  &+Perturbation\\
\midrule
\multirow{4}*{FGSM}     &0.25           &0\%       &0\%      &\multirow{4}*{FGSM}    &0.25           &0.4\%   &0.9\%\\
{}                      &1              &0\%       &0\% 
&{}             &1         &0.9\%   &1.8\%  \\
                        
{}              &8         &0\%       &0\%                 
&{}             &8         &2.6\%   &2.6\%\\
                        
{}             &32             &15.6\%    &16.4\%
&{}            &32             &6.1\%   &8.0\%\\
\midrule

\multirow{4}*{PGD}  &0.25            &0\%       &0\% 
&\multirow{4}*{PGD}  &0.25            &0.4\%    &0.9\%\\
{}                     &1               &0\%       &0\%    
&{}                    &1               &0.8\%    &2.8\% \\
{}                     &8               &22.2\%    &43.4\%
&{}                     &8               &11.5\%   &24.2\%\\
{}                     &32              &33.8\%    &47.2\%
&{}                     &32              &39.0\%   &44.1\%\\

\midrule
\bf{\tool}         &\multicolumn{3}{c|}{\bf{84.4\% }}     &\bf{\tool}         &\multicolumn{3}{c}{\bf{57.7\%}}\\
\bottomrule
\end{tabular}
\end{table*}

\noindent
\textbf{Norm-bounded Adversarial attacks.}
We compare the attack success rate of AdvSPADE with the state-of-the-art norm bounded adversarial attacks, including FGSM and PGD~\cite{43405,madry2018towards}, for two dataset. We set the $l_{\infty}$ norm bound size $\epsilon$ to $\{0.25,1,8,32\}$ for both FGSM and PGD. For PGD, we follow the ~\cite{kurakin2016adversarial,Arnab_2018} and set number of attack iterations to $min\{ \lfloor \epsilon+4 \rfloor,\lceil 1.25\epsilon  \rceil\}$. We apply FGSM and PGD on both real images and synthetic images by vanilla SPADE, and compare their mIoU scores and FID with ours. 
The results (see Table~\ref{Success Rate}) show that PGD and FGSM attacks can barely attack target networks with small bound size ($\epsilon = 0.25,1,8$). 
For instance, FGSM attack with bound size $\epsilon = 1$ on real and vanilla SPADE generated images achieve $0\%$ attack success rate on DRN-105 network on Cityscapes. In contrast, \tool achieves high attack success rate ($84.4\%$ and $57.7\%$ on DRN-105 and Upernet-101, respectively).


\begin{table*}[t]
\caption{\small{\textbf{Attack Effectiveness under white-box setting} for our \tool and norm-bounded attacks (FGSM,PGD) with bound sizes: $0.25,1,8,32$. We show mIoU and FID scores (FID in parentheses) of \tool generated examples and norm-bounded attacks on both real and standard SPADE generated images. 
The results indicate that traditional norm-bounded attacks need large size perturbation to achieve similar mIoU as \tool, which will be easily detectable by the human.  However, our unrestricted adversarial examples remain invisible to humans, which reveals the effectiveness of our proposed method.}
}
\label{norm bounded attack}
\centering
\scriptsize
\setlength{\tabcolsep}{3pt}
\begin{tabular}{lrrr|lrrr}
\toprule
\multicolumn{4}{c|}{DRN-105 (Cityscapes)} & \multicolumn{4}{c}{Upernet-101 (ADE20K)} \\
\midrule
  &Bound Size      &Real Images  &Vanilla SPADE &  &Bound Size     &Real Images  &Vanilla SPADE\\
Method  &($\epsilon$)      &+Perturbation  &+Perturbation &Method  &($\epsilon$)      &+Perturbation  &+Perturbation\\
\midrule
\multirow{4}*{FGSM}     &0.25           &0.557       &0.431 (63.354)      &\multirow{4}*{FGSM}    &0.25           &0.346       &0.286 (33.821)\\
{}                      &1              &0.408       &0.355 (64.455) 
&{}             &1         &0.278       &0.221 (35.254)\\
                        
{}              &8         &0.196       &0.152 (82.144)                 
&{}             &8         &0.178       &0.152 (60.563)\\
                        
{}             &32             &0.009       &0.009 (248.175)
&{}            &32             &0.070       &0.048 (166.724)\\
\midrule

\multirow{4}*{PGD}  &0.25            &0.557       &0.431 (63.354) 
&\multirow{4}*{PGD}  &0.25           &0.346       &0.286 (33.821)\\
{}                     &1            &0.339       &0.287 (63.971)    
&{}                    &1            &0.276       &0.181 (34.876) \\
{}                     &8            &0.036       &0.022 (69.162)
&{}                     &8           &0.070       &0.022 (62.289)\\
{}                     &32           &0.013       &0.009 (89.998)
&{}                     &32          &0.013       &0.007 (113.553)\\

\midrule
\bf{\tool}      &\multicolumn{3}{c|}{\bf{0.01 (67.302) }} &\bf{\tool}     &\multicolumn{3}{c}{\bf{0.011 (53.49)}}\\

\bottomrule
\end{tabular}
\vspace{-5mm}
\end{table*}

\begin{figure*}[h]
\begin{center}
\includegraphics[width=1\linewidth]{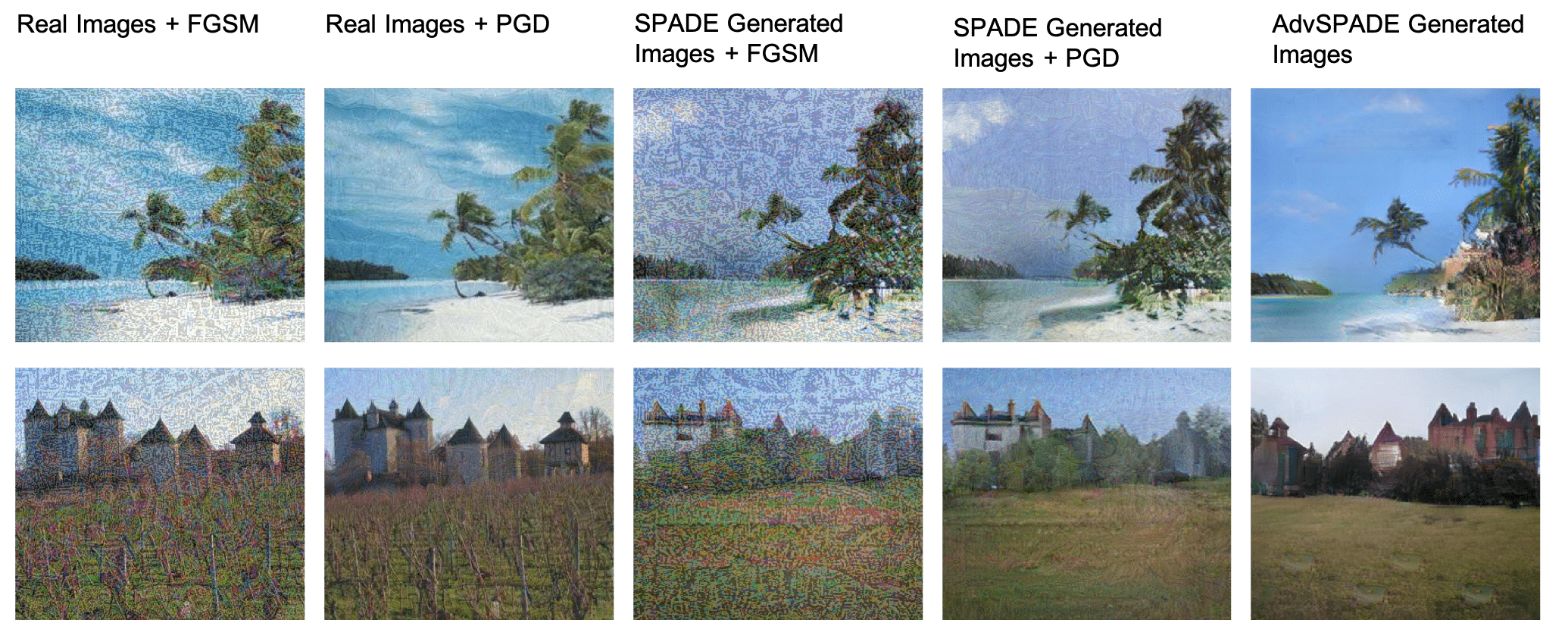}
\end{center}
\caption{\small{\textbf{Comparison of norm-bounded vs.~\tool generated samples at the same mIoU level on ADE20K}. We Apply different attack methods to descend mIoU score to the same level  ($\sim 0.01$) and show the visual comparison. FGSM and PGD are applied with $\epsilon =32$. 
The noise patterns in norm-bounded adversarial images (First four columns) rather than in our examples indicate  that our examples can attack target networks successfully, yet  keep undetectable to human.
}}
\label{fig:fig8}
\vspace{-5mm}
\end{figure*}

Table \ref{norm bounded attack} further reveals that for both FGSM and PGD attack, to decrease the mIoU to the same level as \tool (mIoU = $0.01$), the generated perturbation becomes conspicuous ($\epsilon =32$) so that human can easily distinguish adversarial examples from clean images. FID also reflects the decline of adversarial images' quality. Secondly, adversarial examples generated by FGSM and PGD attack can not make mIoU drop down to the same level as \tool if it is required to maintain the quality of the samples. Consider the adversarial samples on Cityscapes generated by vanilla SPADE and add perturbation with $\epsilon = 1$, their FID (64.455) is comparable with our samples, but mIoU ($0.355$) is much larger than ours ($0.01$). Figure~\ref{fig:fig8} illustrates the difference between AdvSPADE samples and norm-bounded samples on the same level of mIoU score. We can easily see the noise pattern in norm-bounded samples rather than in our examples.

We further compare the transferability of FGSM and PGD with our \tool on black-box setting and observe similar conclusion. The detailed results are shown in supplementary material.

\noindent
\textbf{GAN-based Adversarial Attack.} 
Here, we compare the mIoU drop between Song's unrestricted attack~\cite{song2018constructing} and \tool on Cityscapes (see Table~\ref{mIoU on Cityscapes and ADE20K} ). The results show ~\cite{song2018constructing} generated examples can only lead a small mIoU drop on a white box setting (from 0.62 to 0.461), while \tool generated images drop to 0.010. The transferability of their examples under the blackbox setting is also limited\textemdash mIoU drop of 3\% (Song et al.) vs.~15\% (\tool), demonstrating limitations of their attacks. 

\begin{table}[h]
\caption{\small{\textbf{Comparison between AdvGAN and \tool ours on DRN-105}}. We present the mIoU score and attack success rate of AdvGAN~\cite{Xiao_2018} with two different hyper-parameter setting on Cityscapes. The results show the ineffectiveness of AdvGAN on segmentation task even with large norm-bound size. 
}
\label{advGAN}
\centering
\scriptsize
\begin{tabular}{lrrr}
\toprule
Method        &Bound Size      &Attack Success Rate  &mIoU \\
\midrule
        & 0.25  &0\%       &0.350 \\
AdvGAN  &1      &0\%       &0.344\\
        &8      &0\%       &0.264\\
        &32     &0\%       &0.351\\
\midrule
        & 0.25  &0\%       &0.340\\
AdvGAN  &1      &0\%       &0.338\\
($\lambda_{adv}$ = -100) &8 &7.8\% &0.055\\
        &32     &6.4\%       &0.044\\
\midrule
\bf{AdvSPADE(Ours)}                            &-               &{\bf84.4}\%   &{\bf0.01}\\
\bottomrule
\end{tabular}
\vspace{-5mm}
\end{table}

We also compare the attack success rate and mIoU drop between AdvGAN~\cite{Xiao_2018} and \tool (see Table~\ref{advGAN}). Since AdvGAN can be considered a special norm-bounded attack method, we follow the same experimental setting for bound size with traditional norm-bounded attacks.  
With suggested hyper-parameter setting in~\cite{Xiao_2018}, AdvGAN generated examples show a 0\% attack success rate and only 0.351 mIoU drop even when bound size is as large as 32.  After adjusting the hyper-parameter, 
generated adversarial examples can only attack target segmentation network successfully with large bound size  (Attack success rate is 7.8\%, mIoU is 0.055 at bound size is 8). However, in such a case, we can clearly see the noise pattern from adversarial examples. From our experimental results, it turns out that AdvGAN's attack effectiveness is even worse than the traditional norm-bound attacks, showing that its applicability to only classification tasks.

\begin{table}[h]
\caption{\textbf{\small Ablation Study Results for Cityscapes Dataset}}
\label{Ablation}
\centering
\scriptsize
\begin{tabular}{l|rrr}
\toprule
Method          &FID        &mIoU       & Attack Success Rate \\
\midrule
w/o Feature Matching Loss   &79.76     &0.027      &57.6\% \\
w/o VGG Loss    &80.142           &0.026           &56.0\%     \\
w/o Adv Loss    &62.939     &0.62       &0\%  \\
\tool       &67.302     &0.01       & 84\% \\
\bottomrule
\end{tabular}
\end{table}





Table~\ref{Ablation} further shows an ablation study on Cityscapes Dataset to evaluate effectiveness of each component in \tool.  The results show that both Feature Matching loss and VGG loss terms are important for the quality and effectiveness of adversarial examples generation. Removing any of them causes FID to increase, and attack success rate to drop. Besides, on removing adversarial loss term \tool degenerates to vanilla SPADE, which can only generate benign images. 

\noindent
\textbf{Human Evaluation.}
Using Amazon Mechanical Turk (AMT), we evaluate how a human perceives  generated adversarial images.  A detailed result is presented in the Supplementary part. This is done in two settings:

\noindent
(1) \textit{Semantic Consistency Test}: If the semantic meanings of our adversarial examples are consistent with their respective ground truth labels, humans will segment the examples correctly. However, asking workers to segment every pixel is time-consuming and inefficient. Instead, we give AMT workers a pair of images: a generated adversarial image and a semantic label (half of the images pairs are matched, and rest are mismatched) and ask them if the semantic meaning of given synthetic image is consistent with the given semantic label. We notice that users can identify the semantic meaning of our adversarial examples precisely ($94.4\%$ for Cityscapes, $98.0\%$ for ADE20K). 
This shows although a segmentation network completely fails to handle our adversarial examples, humans can successfully identify the semantic meaning, which proves our attack's success.

\noindent
(2) \textit{Fidelity AB Test}: We compare the visual fidelity of  \tool with vanilla SPADE. We give workers the semantic ground truth label and two generated images by \tool and vanilla SPADE respectively and ask them to select the more appropriate image corresponding to the ground truth label.  $68.4\%$  and $75.2\%$ users favor our examples over vanilla SPADE for Cityscapes and ADE20K dataset indicating competitive visual fidelity of our adversarial images.

\noindent
\textbf{Robustness Evaluation.}
We first show that robust training with norm-bounded adversarial images can defend restricted adversarial attacks where perturbation is added either on real or synthetic images. However, our unrestricted adversarial examples can still attack these robust models successfully~\cite{43405}. We then present the experimental results of a more robust segmentation model built based on our unrestricted examples. We follow the training setting introduced by ~\cite{madry2018towards}: we select PGD as the attack method and set the adversarial training epoch = $100$ on Cityscapes, $50$ on ADE20K, norm-bound size $\epsilon = 8$, attack iteration = $10$, step size = $1$. After the training phase, we use PGD with the same setting to generate norm-bounded perturbation and add it on both real  and synthetic images by vanilla SPADE. 
We find that real and synthesized images with perturbation  make mIoU decrease to ${0.325}$ and ${0.225}$ on robust DRN-105, ${0.239}$ and ${0.197}$ on robust Upernet-101, respectively. In contrast, our adversarial examples can achieve ${0.033}$ mIoU score on robust DRN-105, and ${0.024}$ on robust Upernet-101 indicating that our examples can successfully surpass the robust models trained with norm-bounded adversarial examples. Next, we train a model 50 epochs with our unrestricted adversarial examples on the Cityscapes dataset and then apply PGD to attack. The result shows that PGD attack can only achieve $1.8\%$ attack success rate on DRN-105. Since $l_\infty$ norm-bound examples are unknown for the robust model defended by our samples, the low success rate reflects models gain stronger robustness from adversarial training with AdvSPADE examples.

\section{Conclusion}
This paper explores the existence of adversarial examples beyond norm-bounded metric on the state-of-the-art semantic segmentation neural networks. By modifying the loss function of SPADE architecture, we are able to generate high quality unrestricted realistic adversarial examples, 
which mislead segmentation networks' behavior. We demonstrate the effectiveness and robustness of our method by comparing  with traditional norm-bounded attacks. We also show that our generated adversarial examples can easily surpass the state-of-the-art defense method, which raises new concerns about the security of segmentation networks.


{\small
\bibliographystyle{cvpr2020}
\bibliography{cvpr2020}

\begin{thebibliography}{10}\itemsep=-1pt

\bibitem{Arnab_2018}
Anurag Arnab, Ondrej Miksik, and Philip~H.S. Torr.
\newblock On the robustness of semantic segmentation models to adversarial
  attacks.
\newblock {\em 2018 IEEE/CVF Conference on Computer Vision and Pattern
  Recognition}, Jun 2018.

\bibitem{athalye2018obfuscated}
Anish Athalye, Nicholas Carlini, and David Wagner.
\newblock Obfuscated gradients give a false sense of security: Circumventing
  defenses to adversarial examples, 2018.

\bibitem{badrinarayanan2017segnet}
Vijay Badrinarayanan, Alex Kendall, and Roberto Cipolla.
\newblock Segnet: A deep convolutional encoder-decoder architecture for image
  segmentation.
\newblock {\em IEEE transactions on pattern analysis and machine intelligence},
  39(12):2481--2495, 2017.

\bibitem{Barrow:1981:ILD:3015381.3015385}
H.~G. Barrow and J.~M. Tenenbaum.
\newblock Interpreting line drawings as three-dimensional surfaces.
\newblock {\em Artif. Intell.}, 17(1-3):75--116, Aug. 1981.

\bibitem{CW}
Nicholas Carlini and David~A. Wagner.
\newblock Towards evaluating the robustness of neural networks.
\newblock In {\em 2017 {IEEE} Symposium on Security and Privacy, {SP} 2017, San
  Jose, CA, USA, May 22-26, 2017}, pages 39--57, 2017.

\bibitem{chen2017deeplab}
Liang-Chieh Chen, George Papandreou, Iasonas Kokkinos, Kevin Murphy, and Alan~L
  Yuille.
\newblock Deeplab: Semantic image segmentation with deep convolutional nets,
  atrous convolution, and fully connected crfs.
\newblock {\em IEEE transactions on pattern analysis and machine intelligence},
  40(4):834--848, 2017.

\bibitem{chen2017rethinking}
Liang-Chieh Chen, George Papandreou, Florian Schroff, and Hartwig Adam.
\newblock Rethinking atrous convolution for semantic image segmentation, 2017.

\bibitem{cisse2017houdini}
Moustapha Cisse, Yossi Adi, Natalia Neverova, and Joseph Keshet.
\newblock Houdini: Fooling deep structured prediction models.
\newblock {\em arXiv preprint arXiv:1707.05373}, 2017.

\bibitem{Cordts2016Cityscapes}
Marius Cordts, Mohamed Omran, Sebastian Ramos, Timo Rehfeld, Markus Enzweiler,
  Rodrigo Benenson, Uwe Franke, Stefan Roth, and Bernt Schiele.
\newblock The cityscapes dataset for semantic urban scene understanding.
\newblock In {\em Proc. of the IEEE Conference on Computer Vision and Pattern
  Recognition (CVPR)}, 2016.

\bibitem{dunn2019adaptive}
Isaac Dunn, Hadrien Pouget, Tom Melham, and Daniel Kroening.
\newblock Adaptive generation of unrestricted adversarial inputs, 2019.

\bibitem{engstrom2017exploring}
Logan Engstrom, Brandon Tran, Dimitris Tsipras, Ludwig Schmidt, and Aleksander
  Madry.
\newblock Exploring the landscape of spatial robustness, 2017.

\bibitem{engstrom2017rotation}
Logan Engstrom, Brandon Tran, Dimitris Tsipras, Ludwig Schmidt, and Aleksander
  Madry.
\newblock A rotation and a translation suffice: Fooling cnns with simple
  transformations, 2017.

\bibitem{ess2009segmentation}
Andreas Ess, Tobias Mueller, Helmut Grabner, and Luc~J Van~Gool.
\newblock Segmentation-based urban traffic scene understanding.
\newblock In {\em BMVC}, volume~1, page~2. Citeseer, 2009.

\bibitem{Feinman2017DetectingAS}
Reuben Feinman, Ryan~R. Curtin, Saurabh Shintre, and Andrew~B. Gardner.
\newblock Detecting adversarial samples from artifacts.
\newblock {\em ArXiv}, abs/1703.00410, 2017.

\bibitem{43405}
Ian Goodfellow, Jonathon Shlens, and Christian Szegedy.
\newblock Explaining and harnessing adversarial examples.
\newblock In {\em International Conference on Learning Representations}, 2015.

\bibitem{heusel2017gans}
Martin Heusel, Hubert Ramsauer, Thomas Unterthiner, Bernhard Nessler, and Sepp
  Hochreiter.
\newblock Gans trained by a two time-scale update rule converge to a local nash
  equilibrium, 2017.

\bibitem{44873}
Geoffrey Hinton, Oriol Vinyals, and Jeffrey Dean.
\newblock Distilling the knowledge in a neural network.
\newblock In {\em NIPS Deep Learning and Representation Learning Workshop},
  2015.

\bibitem{hong2018learning}
Seunghoon Hong, Xinchen Yan, Thomas Huang, and Honglak Lee.
\newblock Learning hierarchical semantic image manipulation through structured
  representations, 2018.

\bibitem{ioffe2015batch}
Sergey Ioffe and Christian Szegedy.
\newblock Batch normalization: Accelerating deep network training by reducing
  internal covariate shift, 2015.

\bibitem{Isola_2017}
Phillip Isola, Jun-Yan Zhu, Tinghui Zhou, and Alexei~A. Efros.
\newblock Image-to-image translation with conditional adversarial networks.
\newblock {\em 2017 IEEE Conference on Computer Vision and Pattern Recognition
  (CVPR)}, Jul 2017.

\bibitem{kingma2014adam}
Diederik~P Kingma and Jimmy Ba.
\newblock Adam: A method for stochastic optimization.
\newblock {\em arXiv preprint arXiv:1412.6980}, 2014.

\bibitem{kingma2013autoencoding}
Diederik~P Kingma and Max Welling.
\newblock Auto-encoding variational bayes, 2013.
\newblock cite arxiv:1312.6114.

\bibitem{kurakin2016adversarial}
Alexey Kurakin, Ian Goodfellow, and Samy Bengio.
\newblock Adversarial machine learning at scale, 2016.

\bibitem{NIPS2018_8240}
Donghoon Lee, Sifei Liu, Jinwei Gu, Ming-Yu Liu, Ming-Hsuan Yang, and Jan
  Kautz.
\newblock Context-aware synthesis and placement of object instances.
\newblock In S. Bengio, H. Wallach, H. Larochelle, K. Grauman, N. Cesa-Bianchi,
  and R. Garnett, editors, {\em Advances in Neural Information Processing
  Systems 31}, pages 10393--10403. Curran Associates, Inc., 2018.

\bibitem{long2015fully}
Jonathan Long, Evan Shelhamer, and Trevor Darrell.
\newblock Fully convolutional networks for semantic segmentation.
\newblock In {\em Proceedings of the IEEE conference on computer vision and
  pattern recognition}, pages 3431--3440, 2015.

\bibitem{Lyu_2015}
Chunchuan Lyu, Kaizhu Huang, and Hai-Ning Liang.
\newblock A unified gradient regularization family for adversarial examples.
\newblock {\em 2015 IEEE International Conference on Data Mining}, Nov 2015.

\bibitem{madry2018towards}
Aleksander Madry, Aleksandar Makelov, Ludwig Schmidt, Dimitris Tsipras, and
  Adrian Vladu.
\newblock Towards deep learning models resistant to adversarial attacks.
\newblock In {\em International Conference on Learning Representations}, 2018.

\bibitem{milletari2016v}
Fausto Milletari, Nassir Navab, and Seyed-Ahmad Ahmadi.
\newblock V-net: Fully convolutional neural networks for volumetric medical
  image segmentation.
\newblock In {\em 2016 Fourth International Conference on 3D Vision (3DV)},
  pages 565--571. IEEE, 2016.

\bibitem{mirza2014conditional}
Mehdi Mirza and Simon Osindero.
\newblock Conditional generative adversarial nets, 2014.

\bibitem{miyato2018spectral}
Takeru Miyato, Toshiki Kataoka, Masanori Koyama, and Yuichi Yoshida.
\newblock Spectral normalization for generative adversarial networks.
\newblock {\em arXiv preprint arXiv:1802.05957}, 2018.

\bibitem{Odena2016ConditionalIS}
Augustus Odena, Christopher Olah, and Jonathon Shlens.
\newblock Conditional image synthesis with auxiliary classifier gans.
\newblock In {\em ICML}, 2016.

\bibitem{papernot2017extending}
Nicolas Papernot and Patrick McDaniel.
\newblock Extending defensive distillation, 2017.

\bibitem{papernot2016transferability}
Nicolas Papernot, Patrick McDaniel, and Ian Goodfellow.
\newblock Transferability in machine learning: from phenomena to black-box
  attacks using adversarial samples, 2016.

\bibitem{Papernot_2016}
Nicolas Papernot, Patrick McDaniel, Xi Wu, Somesh Jha, and Ananthram Swami.
\newblock Distillation as a defense to adversarial perturbations against deep
  neural networks.
\newblock {\em 2016 IEEE Symposium on Security and Privacy (SP)}, May 2016.

\bibitem{park2019semantic}
Taesung Park, Ming-Yu Liu, Ting-Chun Wang, and Jun-Yan Zhu.
\newblock Semantic image synthesis with spatially-adaptive normalization, 2019.

\bibitem{Ronneberger_2015}
Olaf Ronneberger, Philipp Fischer, and Thomas Brox.
\newblock U-net: Convolutional networks for biomedical image segmentation.
\newblock {\em Medical Image Computing and Computer-Assisted Intervention –
  MICCAI 2015}, page 234–241, 2015.

\bibitem{Shaham2018UnderstandingAT}
Uri Shaham, Yutaro Yamada, and Sahand Negahban.
\newblock Understanding adversarial training: Increasing local stability of
  supervised models through robust optimization.
\newblock {\em Neurocomputing}, 307:195--204, 2018.

\bibitem{shen2018brain}
Guangyu Shen, Yi Ding, Tian Lan, Hao Chen, and Zhiguang Qin.
\newblock Brain tumor segmentation using concurrent fully convolutional
  networks and conditional random fields.
\newblock In {\em Proceedings of the 3rd International Conference on Multimedia
  and Image Processing}, pages 24--30. ACM, 2018.

\bibitem{song2018constructing}
Yang Song, Rui Shu, Nate Kushman, and Stefano Ermon.
\newblock Constructing unrestricted adversarial examples with generative
  models, 2018.

\bibitem{Sudre_2017}
Carole~H. Sudre, Wenqi Li, Tom Vercauteren, Sebastien Ourselin, and M.
  Jorge~Cardoso.
\newblock Generalised dice overlap as a deep learning loss function for highly
  unbalanced segmentations.
\newblock {\em Lecture Notes in Computer Science}, page 240–248, 2017.

\bibitem{szegedy2013intriguing}
Christian Szegedy, Wojciech Zaremba, Ilya Sutskever, Joan Bruna, Dumitru Erhan,
  Ian Goodfellow, and Rob Fergus.
\newblock Intriguing properties of neural networks, 2013.

\bibitem{tian2018deeptest}
Yuchi Tian, Kexin Pei, Suman Jana, and Baishakhi Ray.
\newblock Deeptest: Automated testing of deep-neural-network-driven autonomous
  cars.
\newblock In {\em Proceedings of the 40th international conference on software
  engineering}, pages 303--314. ACM, 2018.

\bibitem{Wang_2018}
Ting-Chun Wang, Ming-Yu Liu, Jun-Yan Zhu, Andrew Tao, Jan Kautz, and Bryan
  Catanzaro.
\newblock High-resolution image synthesis and semantic manipulation with
  conditional gans.
\newblock {\em 2018 IEEE/CVF Conference on Computer Vision and Pattern
  Recognition}, Jun 2018.

\bibitem{wang2019atgan}
Xiaosen Wang, Kun He, and John~E. Hopcroft.
\newblock At-gan: A generative attack model for adversarial transferring on
  generative adversarial nets, 2019.

\bibitem{wong2019wasserstein}
Eric Wong, Frank~R. Schmidt, and J.~Zico Kolter.
\newblock Wasserstein adversarial examples via projected sinkhorn iterations,
  2019.

\bibitem{Xiao_2018}
Chaowei Xiao, Bo Li, Jun-yan Zhu, Warren He, Mingyan Liu, and Dawn Song.
\newblock Generating adversarial examples with adversarial networks.
\newblock {\em Proceedings of the Twenty-Seventh International Joint Conference
  on Artificial Intelligence}, Jul 2018.

\bibitem{xiao2018unified}
Tete Xiao, Yingcheng Liu, Bolei Zhou, Yuning Jiang, and Jian Sun.
\newblock Unified perceptual parsing for scene understanding.
\newblock In {\em Proceedings of the European Conference on Computer Vision
  (ECCV)}, pages 418--434, 2018.

\bibitem{Xie_2017}
Cihang Xie, Jianyu Wang, Zhishuai Zhang, Yuyin Zhou, Lingxi Xie, and Alan
  Yuille.
\newblock Adversarial examples for semantic segmentation and object detection.
\newblock {\em 2017 IEEE International Conference on Computer Vision (ICCV)},
  Oct 2017.

\bibitem{Xu_2018}
Weilin Xu, David Evans, and Yanjun Qi.
\newblock Feature squeezing: Detecting adversarial examples in deep neural
  networks.
\newblock {\em Proceedings 2018 Network and Distributed System Security
  Symposium}, 2018.

\bibitem{yao20183daware}
Shunyu Yao, Tzu Ming~Harry Hsu, Jun-Yan Zhu, Jiajun Wu, Antonio Torralba,
  William~T. Freeman, and Joshua~B. Tenenbaum.
\newblock 3d-aware scene manipulation via inverse graphics, 2018.

\bibitem{Yu2016}
Fisher Yu and Vladlen Koltun.
\newblock Multi-scale context aggregation by dilated convolutions.
\newblock In {\em International Conference on Learning Representations (ICLR)},
  2016.

\bibitem{yu2017dilated}
Fisher Yu, Vladlen Koltun, and Thomas Funkhouser.
\newblock Dilated residual networks.
\newblock In {\em Proceedings of the IEEE conference on computer vision and
  pattern recognition}, pages 472--480, 2017.

\bibitem{yu2019free}
Jiahui Yu, Zhe Lin, Jimei Yang, Xiaohui Shen, Xin Lu, and Thomas~S Huang.
\newblock Free-form image inpainting with gated convolution.
\newblock In {\em Proceedings of the IEEE International Conference on Computer
  Vision}, pages 4471--4480, 2019.

\bibitem{Zhao_2017}
Hengshuang Zhao, Jianping Shi, Xiaojuan Qi, Xiaogang Wang, and Jiaya Jia.
\newblock Pyramid scene parsing network.
\newblock {\em 2017 IEEE Conference on Computer Vision and Pattern Recognition
  (CVPR)}, Jul 2017.

\bibitem{zhou2016semantic}
Bolei Zhou, Hang Zhao, Xavier Puig, Sanja Fidler, Adela Barriuso, and Antonio
  Torralba.
\newblock Semantic understanding of scenes through the ade20k dataset.
\newblock {\em arXiv preprint arXiv:1608.05442}, 2016.

\bibitem{zhou2018semantic}
Bolei Zhou, Hang Zhao, Xavier Puig, Tete Xiao, Sanja Fidler, Adela Barriuso,
  and Antonio Torralba.
\newblock Semantic understanding of scenes through the ade20k dataset.
\newblock {\em International Journal on Computer Vision}, 2018.

\end{thebibliography}
}

\end{document}